\documentclass[letterpaper, 10 pt, conference]{ieeeconf}  

\IEEEoverridecommandlockouts                              

\usepackage{times}
\usepackage{latexsym}
\usepackage{graphicx}
\usepackage{subcaption}
\usepackage{booktabs}
\usepackage{multirow}
\usepackage{amsmath}
\usepackage{bm}

\title{\LARGE \bf
Self-supervised Document Clustering Based on BERT with Data Augmentation}

\author{Haoxiang Shi$^{1}$, Cen Wang$^{2}$  
\thanks{}
\thanks{$^{1}$Haoxiang Shi and Tetsuya Sakai are with Graduate School of Fundamental Science and Engineering, Waseda University, 169-8050 Tokyo, Japan.
        {\tt\small hollis.shi@toki.waseda.jp} and
        {\tt\small tetsuyasakai@acm.org}}%
\thanks{$^{2}$Cen Wang is with KDDI Research, Inc., 356-8502
        Saitama, Japan.
        {\tt\small ce-wang@kddi-research.jp}}%
}

\begin{document}

\maketitle
\thispagestyle{empty}
\pagestyle{empty}

\begin{abstract}

Contrastive learning is a promising approach to unsupervised learning, as it inherits the advantages of well-studied deep models without a dedicated and complex model design. In this paper, based on bidirectional encoder representations from transformers, we propose self-supervised contrastive learning (SCL) as well as few-shot contrastive learning (FCL) with unsupervised data augmentation (UDA) for text clustering. SCL outperforms state-of-the-art unsupervised clustering approaches for short texts and those for long texts in terms of several clustering evaluation measures. FCL achieves performance close to supervised learning, and FCL with UDA further improves the performance for short texts.

\end{abstract}

\section{INTRODUCTION}

Text clustering is the task of grouping a set of unlabeled texts such that texts in the same cluster are more similar to each other than to those in other clusters. This task is useful for various applications, such as opinion mining, automatic topic labeling~\cite{c1}, language modeling, recommendation~\cite{c2}, and query expansion to improve retrieval~\cite{c3}. Unsupervised learning is a practical approach to clustering, as often there are no ``gold clusters” in real applications. Common unsupervised approaches are generative: they use an encoding network to learn latent representations for input texts and feed the latent representations into another generative network. By minimizing the generation similarity, the most appropriate latent representations can be learned. These generative approaches usually require complex architectures, but are limited in terms of effectiveness.

On the other hand, the recent success of a discriminative contrastive learning (CL) framework in the image classification field~\cite{c4} inspired similar approaches in the NLP field (see Section~\ref{sec:work}). CL can help achieve model training with weak labels or totally without labels. Hence, in the present study, we explore CL-based approaches to tackle the text clustering task. Specifically, based on bidirectional encoder representations from transformers (BERT)~\cite{c5}, we propose self-supervised contrastive learning (SCL) and few-shot contrastive learning (FCL). Our contributions are as follows: 

\begin{itemize}
    \item We propose multi-language \textit{back translation} (BT) and \textit{random masking} (RM) to generate positive samples for SCL.
    \item We propose FCL with unsupervised data augmentation (UDA)~\cite{c6}. 
\end{itemize}

We evaluated the aforementioned learning approaches on two short-text datasets and two long-text datasets, and the results show that SCL achieves state-of-the-art clustering accuracy, and FCL achieves performance close to supervised learning. Moreover, FCL with UDA further improves performance for short texts.

\section{RELATED WORK}
\label{sec:work}

\subsection{Contrastive Learning in NLP}

The recent years have witnessed substantial development in studies related to CL. Gunel et al.~\cite{c7} proposed a supervised CL architecture for multiple language tasks. Fang et al.~\cite{c8} proposed contrastive self-supervised encoder representations from transformers to predict whether two augmented sentences originated from the same sentence. Li et al. ~\cite{c9} proposed CL with mutual information maximization for cross-domain sentiment classification. Wu et al.~\cite{c10} designed a metric that covers both linguistic qualities and semantic informativeness using BERT-based CL. Xiong et al.~\cite{c11} proposed approximate nearest neighbor negative contrastive estimation for dense text retrieval.

\subsection{Text Clustering}
In traditional unsupervised clustering, Yuan et al.~\cite{c12} proposed feature clustering hashing (FCH), and Li et al.~\cite{c13} proposed subspace clustering guided convex non-negative matrix factorization to perform text clustering. Subsequently, Wang et al.~\cite{c14} proposed Gaussian bidirectional adversarial topic (G-BAT) models to achieve higher accuracy among these methods. Autoencoder (AE)~\cite{c15}, graph autoencoder (GAE), and variational graph autoencoder (VGAE)~\cite{c16} can be used for generative unsupervised clustering. Chiu et al.~\cite{c1} provided a summary of comparisons of these methods. Semi-supervised deep clustering methods have also been proposed based on deep architectures, such as deep clustering network (DCN)~\cite{c17} and text convolutional Siamese network~\cite{c18}.

\begin{figure}[ht]
\centering
\includegraphics[scale=0.42]{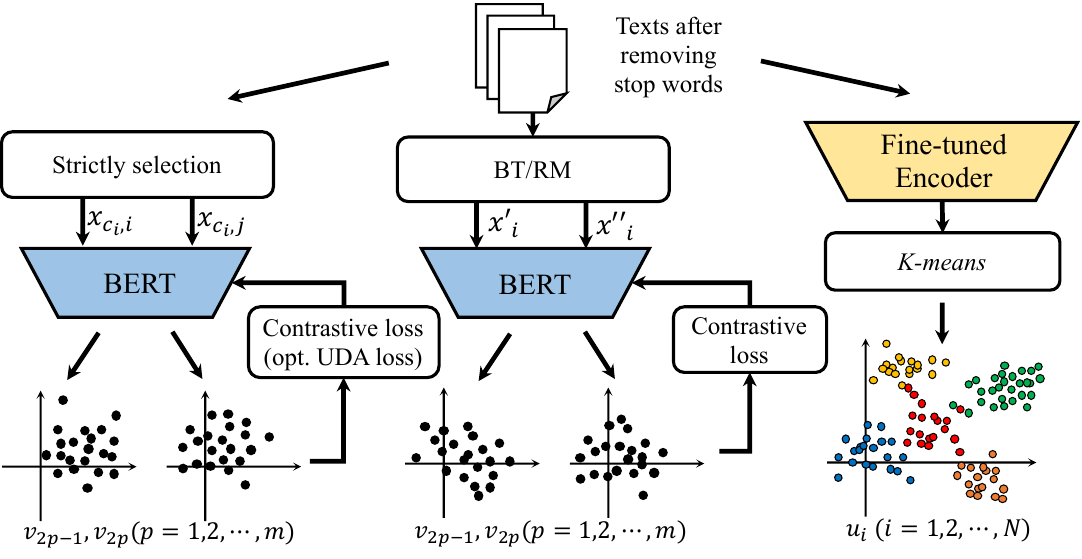}
\caption{Learning framework.}
\label{fig1}
\end{figure}

\section{METHODOLOGY}
\label{sec:method}

Figure~\ref{fig1} illustrates the learning framework for FCL and SCL. During a mini-batch, (1) $m$ pairs of texts are selected/generated (We detail the specific methods of selecting and generating in Section~\ref{mbc}), where intra-pair texts are used as positive samples and inter-pair texts are treated as negative samples; (2) BERT takes each text pair (with stop words removed) as inputs and transforms the texts into the latent representations. (3) BERT is tuned by a contrastive loss, $L_{\rm{CL}}$ (Equation~~\ref{loss}) which is calculated based on the latent representations. After a learning epoch completes, we input all the texts in a dataset into the tuned BERT and obtain the representation $u_i$~$(i=1,2,\cdots,N$, where $N$ is the number of items in the dataset$)$ for clustering.

\subsection{Mini-batch Construction}
\label{mbc}
For SCL, the entire dataset was used for tuning. $m$ texts are first randomly selected from the dataset. We do not require that the texts are from the different classes. For a selected text $x_i, i=1,2,\cdots,m$, $x'_i$ and $x''_i$ are two texts generated by BT (i.e., back translation: given a text in language A, this text is translated into language B, and then is translated back to language A again) using different languages, or by RM (i.e., random masking, to randomly mask some words in a text). The original text is excluded in a mini-batch. Thus, the size of a mini-batch is $2m$. For FCL, BERT is tuned by $m$ pairs of few-shot labeled items in the dataset. The size of a mini-batch is also $2m$. Each pair of (original) texts, $x_{c_{i},i}$ and $x_{c_{i},j}$, are selected from the same class $c_i$, and different pairs are strictly from $n$ different classes. To make full contrasts of the texts in the different classes, we suggest $m \geq n$.

\subsection{Contrastive Loss}
The contrastive loss is the average of the pair losses. In a pair loss $l(i,j)$ (Equation~\ref{lij}), where $i=2p-1$ and $j=2p$ in the $p$-th ($p=1,2,\cdots,m$) pair, the $s_{i, j}$ is the cosine similarity of $v_i$ and $v_j$. The $\tau ( > 0)$ denotes a temperature parameter that can impact the intra-cluster and inter-cluster distance, thereby impacting the clustering accuracy. 

\begin{center}{
	\begin{equation}
		L_{\rm{CL}} = \frac{1}{2m} \sum_{k=1}^{m}[l(2p-1,2p)+l(2p,2p-1)]
		\label{loss}
	\end{equation}
	}
\end{center}

\begin{center}{
	\begin{equation}
		l(i,j) = -\log{\frac{\exp{(s_{i,j}/\tau)}}{\sum_{k=1,k\neq i}^{2m}\exp{(s_{i,k}/\tau)}}}
		\label{lij}
	\end{equation}
	}
\end{center}

\subsection{Unsupervised Data Augmentation}

UDA was originally proposed for emotion analysis~\cite{c6}, which is a binary classification task. When we apply UDA for FCL, every text in a dataset $D$ is back translated to construct $D'$. BERT takes a text $x_i$ in $D$ and three texts $x'_{i,q}(q=1,2,3)$ in $D'$ as inputs, and feeds outputs into a UDA model with parameter set $\theta$ to obtain the distributions, $p_{\theta}(y|x_i)$ and $p_{\theta}(y|x’_{i,q})$. The UDA loss $L_{\rm{UDA}}$ is the average KL-divergence for each pair of $p_{\theta}(y|x_i)$ and $p_{\theta}(y|x’_{i,q})$ in a mini-batch (Equation~\ref{uda}). The model loss shows in Equation~\ref{fl}. 

\begin{center}{
	\begin{equation}
		L_{\rm{UDA}} = \sum_{i=1}^{m} \mathcal{KL}(p_{\theta}(y|x_i) \parallel p_{\theta}(y|x'_{i}))
		\label{uda}
	\end{equation}
	}
\end{center}

\begin{center}{
	\begin{equation}
		\mathcal{L}=L_{\rm{CL}} + L_{\rm{UDA}}
		\label{fl}
	\end{equation}
	}
\end{center}

\section{EXPERIMENT}
\label{sec:exp}

\subsection{Data}
Following previous work in text clustering~\cite{c1, c14, c2, c19}, we evaluated text clustering methods using four text categorization datasets. In these four datasets, SearchSnippets~\cite{c20} and Stackoverflow~\cite{c21} have short texts, whereas 20Newsgroup~\cite{c22} and Reuters~\cite{c23} have long texts. The dataset statistics are in Appendix~\ref{ds}.

\subsection{Settings}
In our main experiments (to perform both SCL and FCL), we used the basic and uncased BERT and $\tau=0.5$ (see Appendix~\ref{t} for details). The dimension of a latent representation $v_i$ is 20. The mini-batch size is $2m=160$ and $2m=40$ for the short texts and the long texts, respectively.  The learning rate is $2\times 10^{-5}$, and the optimizer is \textit{Adam}. 

In SCL, we chose Google Translate to perform English-Spanish-English and English-French-English BT to generate positive pairs. 30\% and 15\% of words are masked for a short text and a long text in RM, respectively. This is because a small percentage setting may cause no words to be masked in a short text, whereas a large percentage may cause a long text to lose contexts. We chose \emph{K-means} as the method to cluster the texts. For FCL as well as FCL with UDA (FCL + UDA), we take 10\% labeled items to tune BERT and cluster all the texts for each dataset.  

\subsection{Evaluation Measures}
For SCL, to enable comparisons with previous works~\cite{c1, c2}, we evaluated clustering methods using clustering accuracy (ACC) and normalized mutual information (NMI) for the short texts, and ACC and adjusted mutual information (AMI)~\cite{c24} for the long texts.  These measures assume that the desired number of clusters $n$ (i.e., the number of gold classes) is given to the system. However, this assumption may not be entirely practical. We therefore additionally adopted BCubed F1 Score~\cite{c25} to evaluate our frameworks while varying the estimated number of gold classes $\hat{n}$. For FCL, we choose ACC, AMI, and adjusted random index (ARI) for comparisons. The definitions and usages of the measures are shown in Appendix \ref{emfc}. 

\begin{table}[ht]
\small
\centering
\begin{tabular}{crr|rr}
\toprule[1pt]
\multirow{2}*{\textbf{Method}} & \multicolumn{2}{c}{SearchSnippets} & \multicolumn{2}{c}{Stackoverflow} \\ 
\cline{2-3} \cline{4-5}
~ & ACC & NMI & ACC & NMI \\
\hline
TF & $24.7^\ast$ & $9.0^\ast$ & $13.5^\ast$ & $7.8^\ast$ \\
TF-IDF & $33.8^\ast$ & $21.4^\ast$ & $20.3^\ast$ & $15.6^\ast$ \\
Skip-Thought & $33.6^\ast$ & $13.8^\ast$ & $9.3^\ast$ & $2.7^\ast$ \\
SIF & $53.4^\ast$ & $36.9^\ast$ & $30.5^\ast$ & $28.9^\ast$ \\
STC$^{2}$ & $77.0^\ast$ & $62.9^\ast$ & $51.1^\ast$ & $49.0^\ast$ \\
SIF + Aut.,Self-Train. & $77.1^\ast$ & $56.7^\ast$ & $59.8^\ast$ & $54.8^\ast$ \\
\hline
SCL (BT) & $\bf{78.2}$ & $63.6$ & $\bf{77.7}$ & $\bf{72.5}$ \\
SCL (RM) & $77.2$ & $\bf{63.8}$ & $67.5$ & $57.4$ \\
\bottomrule[1pt]
\end{tabular}
\caption{\label{short-text} SCL performance comparisons for short-text datasets: $\ast$ denotes the results reported by~\cite{c2}; the \textbf{bold} results are the best scores.}
\end{table}

\begin{table}[ht]
\small
\centering
\begin{tabular}{crr|rr}
\toprule[1pt]
\multirow{2}*{\textbf{Method}} & \multicolumn{2}{c}{20Newsgroup} & \multicolumn{2}{c}{Reuters} \\ 
\cline{2-3} \cline{4-5}
~ & ACC & AMI & ACC & AMI \\
\hline
NMF & $31.9^\star$ & $45.3^\star$ & $49.6^\star$ & $43.8^\star$ \\
TF-IDF & $33.7^\star$ & $41.7^\star$ & $35.0^\star$ & $45.6^\star$ \\
FCH & $33.8^\dagger$ & ~ & ~ & ~ \\
LDA & $37.2^\star$ & $28.8^\star$ & $54.9^\star$ & $50.3^\star$ \\
G-BAT & $41.3^\ddagger$ & ~ & ~ & ~ \\
GAE & $41.4^\star$ & $49.3^\star$ & $52.3^\star$ & $48.4^\star$ \\
AE & $41.7^\star$ & $48.7^\star$ & $53.7^\star$ & $55.0^\star$ \\
VGAE & $43.1^\star$ & $48.1^\star$ & $53.1^\star$ & $53.3^\star$ \\
DCN & $44.0^\ast$ & ~ & ~ & ~ \\
SS-SB-MT & $47.4^\star$ & $\bf{53.0}^\star$ & $56.3^\star$ & $\bf{58.4}^\star$ \\
\hline
SCL (BT) & $\bf{50.1}$ & $47.1$ & $61.7$ & $45.4$\\
SCL (RM) & $44.1$ & $42.2$ & $\bf{64.4}$ & $46.3$ \\
\bottomrule[1pt]
\end{tabular}
\caption{\label{long-text} SCL performance comparisons for long-text datasets: $\star$, $\dagger$, $\ddagger$, and $\ast$ denote the results reported by~\cite{c1},~\cite{c12}, ~\cite{c14}, and~\cite{c19} respectively; the \textbf{bold} results are the best scores. }

\end{table}

\begin{table}[ht]
\small
\centering
\begin{tabular}{ccr|r|r|r}
\toprule[1pt]
\textbf{Method} & $\bm{n}$ & SearchS. & Stack. & 20N.G. & Reut.\\ 
\hline
~ & $5$ & $44.8$ & $27.2$ & $25.7$ & $\bf{48.9}$\\
SCL & $10$ & $\bf{46.2}$ & $43.0$ & $34.3$ & $47.4$ \\
(BT) & $15$ & $37.6$ & $55.9$ & $\bf{36.0}$ & $40.4$ \\
~ & $20$ & $31.0$ & $\bf{64.5}$ & $34.7$ & $33.4$ \\
\hline
~ & $5$ & $49.1$ & $26.2$ & $23.2$ & $\bf{53.5}$ \\
SCL & $10$ & $\bf{49.3}$ & $38.8$ & $30.2$ & $51.5$ \\
(RM) & $15$ & $40.6$ & $47.1$ & $\bf{31.5}$ & $45.6$ \\
~ & $20$ & $34.03$ & $\bf{51.3}$ & $30.9$ & $35.0$ \\
\bottomrule[1pt]
\end{tabular}
\caption{\label{f1} BCubed F1 Scores under SCL for the  datasets with unknown $n$ from gold clusters; the \textbf{bold} results are the best scores.}
\end{table}

\begin{table*}[ht]
\small
\centering
\begin{tabular}{crrr|rrr|rrr|rrr}
\toprule[1pt]
\multirow{2}*{\textbf{Method}} & \multicolumn{3}{c}{SearchSnippets} & \multicolumn{3}{c}{Stackoverflow} & \multicolumn{3}{c}{20Newsgroup} & \multicolumn{3}{c}{Reuters} \\ 
\cline{2-4} \cline{5-7} \cline{8-10} \cline{11-13}
~ & ACC & AMI & ARI & ACC & AMI & ARI & ACC & AMI & ARI & ACC & AMI & ARI \\
\hline
10\%FCL & $89.6$ & $75.9$ & $77.4$ & $86.0$ & $79.2$ & $79.1$ & $\bf{66.0}$ & $\bf{59.5}$ & $\bf{59.5}$ & $71.3$ & $\bf{62.0}$ & $\bf{67.9}$\\
10\%FCL + UDA & $\bf{90.7}$ & $\bf{78.1}$ & $\bf{79.7}$ & $\bf{87.2}$ & $\bf{80.6}$ & $\bf{80.7}$ & $62.1$ & $54.6$ & $54.0$ & $\bf{78.1}$ & $58.5$ & $58.5$\\ 
\hline
Supervised (ref.) & $96.0$ & $88.7$ & $90.5$ & $89.0$ & $81.0$ & $78.2$ & $69.1$ & $58.8$ & $49.7$ & $88.5$ & $70.0$ & $80.7$ \\
\bottomrule[1pt]
\end{tabular}
\caption{\label{fcl} FCL performance comparisons: the \textbf{bold} results are the best scores. }
\end{table*}

\subsection{Baselines}
\label{base}
For short texts, we compared SCL with SIF + Aut.,Self-Train. (i.e., the state-of-the-art model) and the other methods reported by Hadifar et al~\cite{c2}. For long texts, we primarily compared SCL with SS-SB-MT~\cite{c1} (i.e., the state-of-the-art model) and G-BAT~\cite{c14}. We also included the other methods such as GAE, AE and VAGE reported by Chiu et al~\cite{c1}. Simple illustrations of these baselines are presented in Appendix~\ref{bl}.

We compared FCL with a supervised method that takes training data and cross-entropy loss to tune BERT~\cite{c26}. Except for 20Newsgroup which has the training data (which occupy approximately 70\% of the total data), for the datasets which are not originally divided into training and test parts, we randomly select 70\% of the items in each dataset to tune the models, and the other 30\% for clustering evaluation.

\subsection{Results}
 
\textbf{SCL performance.} In Table~\ref{short-text} (results for short texts), for SearchSnippets, SCL (BT) outperforms SIF + Aut.,Self-Train. by 1.1 points in terms of ACC, and SCL (RM) outperforms that by 7.1 points in terms of NMI. In contrast, for Stackoverflow, our results show greater improvements. More specifically, SCL (BT) outperforms SIF + Aut.,Self-Train. by 17.9 and 17.7 points in ACC and NMI, respectively. As shown in Table~\ref{long-text} (results for long texts), for 20Newsgroup, SCL (BT) outperforms SS-SB-MT and G-BAT by 2.7 points and 8.8 points in ACC, respectively; meanwhile, for Reuters, both SCL (BT) and SCL (RM) outperform SS-SB-MT and G-BAT in ACC. Furthermore, SCL (RM) achieves the best ACC. However, SS-SB-MT outperforms SCL in AMI for 20Newsgroup and Reuters. This may be because SS-SB-MT uses topics to build graphs for the texts, and the clustering results are more interpretable. We further compare BCubed F1 Score for SCL (BT) and SCL (RM) in Table~\ref{f1} under different numbers of clusters, $\hat{n}$. For SearchSnippets and Stackoverflow, the best scores are obtained when $\hat{n}$ is close to the $n$ of the gold clusters (8 and 20, respectively). However, for 20Newsgroup and Reuters, we obtained the best scores when $\hat{n}$ are $15$ and $5$, respectively.  

\textbf{FCL Performance.} Comparing FCL (a semi-supervised model) to SCL and other unsupervised methods may not be fair. Therefore, we compare the FCL results to the aforementioned supervised method (see Section~\ref{base}). In Table~\ref{fcl}, FCL underperforms the supervised method by 6.4 points and 3.0 points for SearchSnippets and Stackoverflow in ACC, respectively. When UDA is added in addition (FCL + UDA), we can obtain higher ACC and AMI scores for both datasets. FCL underperforms the supervised method by 3.1 points and 17.2 points in ACC for 20Newsgroup and Reuters, respectively. However, for FCL + UDA, this difference reduces to 10.4 points for Reuters.

\textbf{Discussion on the fairness.} As SCL is an unsupervised method, we compared it with state-of-the-art unsupervised methods. However, we acknowledge that SCL exploits BT and RM as an additional knowledge source. The main disadvantage of SCL is that, in a mini-batch, two positive pairs may be generated from two texts in the same class. However, we treat these two pairs as negative samples for each other, which limits the performance of SCL. As for FCL, we compared it with a supervised method rather than other few-shot learning methods. Note that while FCL is semi-supervised, it is different from traditional few-short learning methods. More specifically, although we use the texts for all classes when constructing a mini-batch, the only prior knowledge needed is whether the two texts are similar or not. That is, we can transfer an $n$-clustering problem using $n$ classes of labels to one using binary discriminative labels.

\section{CONCLUSION}
In this paper, we proposed SCL and FCL with UDA for text clustering. We tuned BERT by our learning methods, and used the learned latent representations to perform clustering. In SCL, we introduced two data augmentation methods, back translation and random masking. Our experimental results show that SCL achieves the best ACC for all the datasets. In particular, the SCL outperforms the state-of-the-art in terms of both ACC and NMI for short-text clustering. To the best of our knowledge, we are the first to apply CL in an unsupervised manner to perform NLP tasks. The results also show that FCL can obtain performance very close to a supervised method, and FCL with UDA appears to further improve the performance for short texts.




\appendix
\label{sec:app}
\subsection{Dataset Statistics}
\label{ds}
The dataset statistics are shown in Table~\ref{dataset}, where $N$ is the total number of texts in a dataset, $T$ is the total number of tokens, $L_{Avg.}$ is the average length of the texts, and $n$ is the number of the gold clusters. As for Reuters, due to the imbalance in the amount of texts in different classes, we choose the 10 largest classes.

\begin{table}[h!]
\small
\centering
\begin{tabular}{cllll}
\toprule[1pt]
\textbf{Dataset} & $N$ & $T$ & $L_{Avg.}$ & $n$ \\ 
\hline
SearchSnippets & $12.3$k & $31$k & $17.9$ & $8$ \\
Stackoverflow & $20$k & $23$k & $8.3$ & $20$ \\
\hline
20Newsgroup & $1.8$k & $56$k & $245$ & $20$\\
Reuters & $7.6$k & $28$k & $141$ & $10$\\
\bottomrule[1pt]
\end{tabular}
\caption{\label{dataset} Datasets statistics.}
\end{table} 

\subsection{Tuning}
\label{t}
\textbf{Tuning batch size.} The batch size in our experiment was small (i.e., $2m=160$ for short texts and $2m=40$ for long texts) due to resource (i.e., GPUs) constraints, however, according to~\cite{c4} who uses TPUs, we think we can improve performance given a larger batch size.

\textbf{Tuning $\tau$.} To examine the effect of the choice of the temperature parameter $\tau$ on clustering accuracy, we report on the results of an additional experiment using different $\tau$ values. Note that the purpose of this experiment is to investigate the best possible performance; as we are exploiting gold clusters to compute cluster accuracy, tuning $\tau$s in this manner is not practical. From the clustering accuracies for the 20Newsgroup under $\tau$ variations shown in Table~\ref{tau}, it can be seen that both 10\% FCL with UDA and SCL with BT can achieve the best performance when $\tau$ is 0.5, and the accuracy decreases neither when $\tau < 0.5$ or when $\tau > 0.5$. Therefore, we suggest setting $\tau = 0.5$ as the default for the clustering tasks.

\begin{table}[h!]
\small
\centering
\begin{tabular}{lrrrr}
\toprule[1pt]
\textbf{Method~~~$\tau=$} & $0.25$ & $0.5$ & $0.75$ & $1$ \\ 
\hline
SCL (BT) & $45.3$ & $50.1$ & $49.6$ & $44.6$ \\
10\%FCL + UDA & $61.0$ & $62.1$ & $56.4$ & $50.5$ \\
\bottomrule[1pt]
\end{tabular}
\caption{\label{tau} Clustering ACC for 20Newsgroup under $\tau$ variations.}
\end{table}

\subsection{Evaluation Measures for Comparisons}
\label{emfc}

\textbf{ACC.} The ACC requires the mapping between the predicted clusters and the gold clusters, which is defined as Equation~\ref{acc}:
\begin{center}
	\begin{equation}
	ACC = {\sum_{i=1}^{n}\delta(c_{{t}_{i}},map(\hat{c}_{{t}_{i}})) \over N}
	\label{acc}
	\end{equation}
\end{center}
where $n$ is the number of the clusters, and $N$ is the total number of texts. $c_{t_{i}}$ is the ground truth cluster of text $i$, and  ${\hat{c}}_{t_{i}}$ is the predicted cluster of text $i$. If $c_{t_{i}}=map({\hat{c}}_{t_{i}})$, then $\delta(c_{t_{i}}, map({\hat{c}}_{t_{i}}))=1$. The function $map(\cdot)$ indicates a permutation mapping that best matches the predicted clusters to the ground truth classes. 

The method to calculate clustering accuracy follows CoClust~\cite{c27}. First, we construct a $L \times L$ matrix, where $L=\max{(|C|,|\hat{C}|)}$, and $C$ and $\hat{C}$ are the ground true partition (i.e., the gold clusters) and the predicted partition of a dataset. This measure assumes a previously known $|C|=n$, thus $|\hat{C}|=|C|=n$. Then, we count the number of repeated texts in $C_i$ and ${\hat{C}}_j$ as $L(i,j)$, where $C_i$ or ${\hat{C}}_j$ is the $i$-th class or $j$-th cluster of $C$ or $\hat{C}$, respectively. $L$ now can be seen as an adjacent matrix of a bi-graph, and we can applies the method such as the \emph{Hungarian} algorithm to find the maximum perfect match of this bi-graph. Thus, we can get the best mapping between $C$ and $\hat{C}$. We can calculate ACC and other metrics based on this mapping.

\textbf{NMI.} Denote $MI(C, \hat{C})$ as the mutual information:
\begin{center}
	\begin{equation}
	MI(C,\hat{C}) = \sum_{i=1}^{|C|}\sum_{j=1}^{|\hat{C}|} P_{C\hat{C}}(i,j)\log{\frac{P_{C\hat{C}}(i,j)}{P_C(i)P_{\hat{C}}(j)}}
	\label{mi}
	\end{equation}
\end{center}
\noindent where $P_C(i)$ and $P_{\hat{C}}(j)$ denote the probabilities of $i$ in $C$ and $j$ in $\hat{C}$, respectively. $P_{C\hat{C}}(i,j)$ then denotes the joint probability of $i$ in $C$ and $j$ in $\hat{C}$.

Then NMI is defined as:
\begin{center}
	\begin{equation}
		NMI=\frac{MI(C,\hat{C})}{H(C)+H(\hat{C})}
		\label{nmi}
	\end{equation}
\end{center}
\noindent where $H(C)$ and $H(\hat{C})$ are the entropy of $C$ and $\hat{C}$, respectively. 

\textbf{AMI.} To further consider the randomness of $\hat{C}$, AMI introduce the expected MI between $C$ and $\hat{C}$ to adjust the chance:
\begin{center}
	\begin{equation}
	    \begin{split}
		&E\{MI(C,\hat{C})\}= \\ 
		&\sum_{i=1}^{|C|}\sum_{j=1}^{|\hat{C}|}\sum_{n_{ij}=(a_i+b_j-N)^+} \frac{n_{ij}}{N} \log{(\frac{N\cdot n_{ij}}{a_i b_j})} \times \\
		&\frac{a_i!b_j!(N-a_i)!(N-b_j)!}{N!n_{ij}!(a_i-n_{ij})!(b_j-n_{ij})!(N-a_i-b_j+n_{ij})!}
		\end{split}
		\label{emi}
	\end{equation}
\end{center}
\noindent where $n_{ij}=|C_i \cap \hat{C}_j|$, $a_i=\sum_{j=1}^{|\hat{C}|} n_{ij}$, $b_j=\sum_{i=1}^{|C|} n_{ij}$, and $(a_i+b_j-N)^+$ denotes $\max(1, a_i+b_j-N)$. The adjusted measure for the mutual information is defined to be:
\begin{center}
	\begin{equation}
	\begin{split}
	&AMI(C,\hat{C}) = \\ 
	&\frac{MI(C,\hat{C}) - E\{MI(C,\hat{C})\}}{\max \{H(C),H(\hat{C})\} - E\{MI(C,\hat{C})\}}
	\label{ami}
	\end{split}
	\end{equation}
\end{center}

\noindent AMI takes a value of 1 when the two partitions are identical and 0 when the MI between two partitions equals the value expected due to chance alone.

\textbf{ARI.} ARI is the corrected-for-chance version of the Rand Index, which can also measure the similarity between $C$ and $\hat{C}$ when there are no labels. ARI is defined as:
  \begin{equation}
    \begin{split}
    	&ARI=\frac{\sum_{i}\sum_{j}\tbinom{n_{ij}}{2}- \frac{\sum_{i}\tbinom{a_i}{2}\sum_{j}\tbinom{b_j}{2}}{\tbinom{N}{2}}}
    	{\frac{1}{2} \left[\sum_{i}\tbinom{a_i}{2} + \sum_{j}\tbinom{b_j}{2} \right] - \frac{\sum_{i}\tbinom{a_i}{2}\sum_{j}\tbinom{b_j}{2}}{\tbinom{N}{2}}}
    \end{split}
 	\label{ari}
  \end{equation}
 
\noindent where $\tbinom{x}{y}$ denotes the number of combinations when selecting $y$ items from $x$. $n_{ij}$, $a_i$ and $b_j$ have the same meanings with those in Equation \ref{emi}.

\textbf{BCubed F1 Score.} The precision of an item $i$ in the dataset is like:
\begin{center}
	\begin{equation}
		p(i) = \frac{N_{\hat{C}(i)\cap C(i)}}{N_{\hat{C}(i)}}
	\end{equation}
\end{center}where $\hat{C}(i)$ is the predicted cluster that $i$ belongs to, and $C(i)$ is the gold cluster that $i$ belongs to. $N_{{\hat{C}(i)}\cap{C(i)}}$ is the number of items both in $\hat{C}(i)$ and in $C(i)$. $N_{\hat{C}(i)}$ is the number of items in $\hat{C}(i)$.

Then, the recall of $i$ is defined as:
\begin{center}
	\begin{equation}
		r(i) = \frac{N_{\hat{C}(i)\cap C(i)}}{N_{C(i)}}
	\end{equation}
\end{center} where $C(i)$ is the number of items in $C(i)$. 

Then, the score for $i$ is like:
\begin{center}
	\begin{equation}
		f(i) = (1+\beta^2)\frac{p(i) \cdot r(i)}{\beta^2 p(i) + r(i)}
	\end{equation}
\end{center}

\noindent Finally, the Bcubed Precision $P$, Recall $R$ and $\rm{F}_{\beta}$ Score $F$ for a dataset with $N$ items are as follows:
\begin{center}
	\begin{equation}
		P = \frac{1}{N} \sum_{i}p(i)
	\end{equation}
\end{center}

\begin{center}
	\begin{equation}
		R = \frac{1}{N} \sum_{i}r(i)
	\end{equation}
\end{center}

\begin{center}
	\begin{equation}
		F = \frac{1}{N} \sum_{i}f(i)
	\end{equation}
\end{center}

\subsection{Baselines}
\label{bl}
\textbf{SIF + Aut.,Self-Train.} The model includes three steps: (1) short texts are embedded using Smooth Inverse Frequency (SIF) embeddings; (2) during a pre-training phase, a deep autoencoder is applied to encode and reconstruct the short-text SIF embeddings; (3) in a self-training phase, we use soft cluster assignments as an auxiliary target distribution, and jointly fine-tune the encoder weights and the clustering assignments.

\textbf{G-BAT.} The proposed bidirectional adversarial training (BAT) consists of three components: (1) the Encoder $E$ takes the $V$-dimensional document representation $d_r$ sampled from text corpus $C$ as input and transforms it into the corresponding $K$-dimensional topic distribution $\theta_r$; (2) the Generator $G$ takes a random topic distribution $\theta_f$ drawn from a Dirichlet prior as input and generates a $V$-dimensional fake word distribution $d_f$ ; (3) the Discriminator $D$ takes the real distribution pair $p_r=[\theta_r;d_r]$ and fake distribution pair $p_f=[\theta_f;d_f]$ as input and discriminates the real distribution pairs from the fake ones. The outputs of the discriminator are used as supervision signals to learn $E$, $G$ and $D$ during adversarial training. 

In BAT, the generator models topics based on the bag-of-words assumption as in most other neural topic models. To incorporate the word relatedness information captured in word embeddings into the inference process, we modify the generator of BAT and propose Gaussian-BAT, in which $G$ models each topic with a multivariate Gaussian.

\textbf{SS-SB-MT}. This method builds a keyword correlation graph (KCG) for a text using node features (embeddings from a SBERT), word co-occurrence edges, sentence similarity edges and sentence position edges. Then the constructed graphs of the texts are fed into a Multi-Task GAE (MTGAE). Clustering is based on the latent representations from MTGAE. The name SS-SB-MT comes from \textbf{S}entence \textbf{S}imilarity, \textbf{SB}ERT and \textbf{MT}GAE.

\section*{ACKNOWLEDGMENT}
We sincerely thank Dr. Zhaohao Zeng for the paper reviewing and Miss. Jing Shen for the provision of the GPU.


\begin{thebibliography}{99}

\bibitem{c1} Billy Chiu, Sunil Kumar Sahu, Derek Thomas, Neha Sengupta and Mohammady Mahdy, ``Autoencoding keyword correlation graph for document clustering," in Proceedings of the 58th Annual Meeting of the Association for Computational Linguistics, 2020, pp. 3974–3981.
\bibitem{c2} Amir Hadifar, Lucas Sterckx, Thomas Demeester and Chris Develder, ``A self-training approach for short text clustering," in Proceedings of the 4th Workshop on Representation Learning for NLP (RepL4NLP-2019), 2019, pp. 194–199.
\bibitem{c3} Charu C. Aggarwal and ChengXiang Zhai, ``A survey of text clustering algorithms," In Mining text data", Springer, 2012, pp. 77–128.
\bibitem{c4} Ting Chen, Simon Kornblith, Mohammad Norouzi and Geoffrey Hinton, ``A simple framework for contrastive learning of visual representations," arXiv preprint, 2020, arXiv:2002.05709.
\bibitem{c5} Jacob Devlin, Ming-Wei Chang, Kenton Lee and Kristina Toutanova, ``Bert: Pre-training of deep bidirectional transformers for language understanding," arXiv preprint, 2018, arXiv:1810.04805.
\bibitem{c6} Qizhe Xie, Zihang Dai, Eduard Hovy, Minh-Thang Luong and Quoc V Le, ``Unsupervised data augmentation for consistency training," arXiv preprint, 2019, arXiv:1904.12848.
\bibitem{c7} Beliz Gunel, Jingfei Du, Alexis Conneau and Ves Stoyanov, ``Supervised contrastive learning for pre-trained language model fine-tuning," arXiv preprint, 2020, arXiv:2011.01403.
\bibitem{c8} Hongchao Fang and Pengtao Xie, ``Cert: Contrastive self-supervised learning for language understanding," arXiv preprint, 2020, arXiv:2005.12766.
\bibitem{c9} Tian Li, Xiang Chen, Shanghang Zhang, Zhen Dong and Kurt Keutzer, ``Cross-domain sentiment classification with contrastive learning and mutual information maximization," arXiv preprint, 2020, arXiv:2010.16088.
\bibitem{c10} Hanlu Wu, Tengfei Ma, Lingfei Wu, Tariro Manyumwa and Shouling Ji, ``Unsupervised 485 reference-free summary quality evaluation via contrastive learning," arXiv preprint, 2020, arXiv:2010.01781.
\bibitem{c11} Lee Xiong, Chenyan Xiong, Ye Li, Kwok-Fung Tang, Jialin Liu, Paul Bennett, Junaid Ahmed and Arnold Overwijk, ``Approximate nearest neighbor negative contrastive learning for dense text retrieval," arXiv preprint, 2020, arXiv:2007.00808.
\bibitem{c12} Tongtong Yuan, Weihong Deng, Jiani Hu, Zhanfu An and Yinan Tang, ``Unsupervised adaptive hashing based on feature clustering," Neurocomputing, vol. 323, 2019, pp. 373–382.
\bibitem{c13} Xiaocui Li, Hongzhi Yin, Ke Zhou and Xiaofang Zhou, ``Semi-supervised clustering with deep metric learning and graph embedding," World Wide Web, vol. 23, no. 2, 2020, pp. 781–798.
\bibitem{c14} Rui Wang, Xuemeng Hu, Deyu Zhou, Yulan He, Yuxuan Xiong, Chenchen Ye and Haiyang Xu, ``Neural topic modeling with bidirectional adversarial training," arXiv preprint, 2020, arXiv:2004.12331.
\bibitem{c15} Geoffrey E Hinton and Ruslan R Salakhutdinov, ``Reducing the dimensionality of data with neural networks," Science, vol. 313, no. 5786, 2019, pp. 504–507.
\bibitem{c16} Thomas N Kipf and Max Welling, ``Variational graph auto-encoders," arXiv preprint, 2016, arXiv:1611.07308.
\bibitem{c17} Ankita Shukla, Gullal S Cheema and Saket Anand, ``Semi-supervised clustering with neural networks," in 2020 IEEE Sixth International Conference on Multimedia Big Data (BigMM), 2020, IEEE, pp. 152–465.
\bibitem{c18} Lucas Akayama Vilhagra, Eraldo Rezende Fernandes and Bruno Magalhaes Nogueira, ``Textcsn: a semi-supervised approach for text clustering using pairwise constraints and convolutional siamese network," in Proceedings of the 35th Annual ACM Symposium on Applied Computing, 2020, pp. 1135–1142.
\bibitem{c19} Bo Yang, Xiao Fu, Nicholas D Sidiropoulos and Mingyi Hong, ``Towards k-means-friendly spaces: simultaneous deep learning and clustering," in International Conference on Machine Learning, 2017, PMLR, pp. 3861–3870.
\bibitem{c20} Xuan Hieu Phan, Minh Le Nguyen and Susumu Horiguchi, ``Learning to classify short and sparse text web with hidden topics from large-scale data collections," in Proceedings of the 17th International Conference on World Wide Web, Beijing, China, April 21-25, 2008. 
\bibitem{c21} Jiaming Xu, Bo Xu, Peng Wang, Suncong Zheng, Guanhua Tian and Jun Zhao, ``Self-taught con- 497 volutional neural networks for short text clustering," Neural Networks, vol. 88, 2017, pp. 22–31.
\bibitem{c22} Ken Lang, ``Newsweeder: Learning to filter netnews," in Machine Learning Proceedings,  1995, Elsevier, pp. 331–339.
\bibitem{c23} David D. Lewis, Yiming Yang, Tony G. Rose and Fan Li, ``Rcv1: A new benchmark collection for text categorization research," Journal of Machine Learning Research, 2004, vol. 5, no. 2, pp. 361–397.
\bibitem{c24} Nguyen Xuan Vinh, Julien Epps and James Bailey, ``Information theoretic measures for clusterings comparison: Variants, properties, normalization and correction for chance," the Journal of Machine Learning Research, vol. 11, 2010, pp. 2837–2854.
\bibitem{c25} Yanan Qian, Qinghua Zheng, Tetsuya Sakai, Junting Ye and Jun Liu, ``Dynamic author name disambiguation for growing digital libraries," Information Retrieval Journal, vol. 18, no. 5, 2015, pp. 379–412.
\bibitem{c26} Zhiguo Wang, Haitao Mi, and Abraham Ittycheriah, ``Semi-supervised clustering for short text via deep representation learning," arXiv preprint, 2016, arXiv:1602.06797.
\bibitem{c27} Francois Role, Stanislas Morbieu and Mohamed Nadif, ``Coclust: a python package for co-clustering," Journal of Statistical Software, vol. 88, no. 7, 2018, pp. 1–29.








\end{thebibliography}
\end{document}